\documentclass{bmvc2k}
\usepackage{amsmath}
\usepackage{amssymb}

\usepackage{stackengine} 
\usepackage{dblfloatfix} 
\usepackage{pifont}
\usepackage{mathtools}
\newcommand\oast{\stackMath\mathbin{\stackinset{c}{0ex}{c}{0ex}{\ast}{\bigcirc}}}
\newcommand{\norm}[1]{\left\lVert#1\right\rVert}

\newcommand{\mytensor}[1]{\ensuremath{\mathcal{#1}}}
\newcommand{\myvector}[1]{\ensuremath{\mathbf{#1}}}

\newcommand{\mycolon}{\,\boldsymbol{:}\,}

\usepackage[normalem]{ulem}

\title{XNOR-Net++: Improved binary neural networks}

\addauthor{Adrian Bulat}{adrian.bulat@samsung.com}{1}
\addauthor{Georgios Tzimiropoulos}{georgios.t@samsung.com}{1}

\addinstitution{
 Samsung AI Center\\
 Cambridge, UK
}

\runninghead{Bulat, Tzimiropoulos}{XNOR-Net++: Improved Binary Neural Networks}

\begin{document}

\maketitle

\begin{abstract}
This paper proposes an improved training algorithm for binary neural networks in which both weights and activations are binary numbers. A key but fairly overlooked feature of the current state-of-the-art method of XNOR-Net~\cite{rastegari2016xnor} is the use of analytically calculated real-valued scaling factors for re-weighting the output of binary convolutions. We argue that analytic calculation of these factors is sub-optimal. Instead, in this work, we make the following contributions: (a) we propose to fuse the activation and weight scaling factors into a single one that is learned discriminatively via backpropagation. (b) More importantly, we explore several ways of constructing the shape of the scale factors while keeping the computational budget fixed. (c) We empirically measure the accuracy of our approximations and show that they are significantly more accurate than the analytically calculated one. (d) We show that our approach significantly outperforms XNOR-Net within the same computational budget when tested on the challenging task of ImageNet classification, offering up to 6\% accuracy gain.  
\end{abstract}

\section{Introduction}\label{sec:intro}

An open problem in deep learning is how to port recent developments into devices other than desktop machines with one or more high-end GPUs, such as the devices that billions of users use in their everyday life and work like cars, smart-phones, tablets, TVs etc. The straightforward approach to solving this problem is to train models that are both smaller and faster. One of the prominent methods for achieving both goals is through training binary networks, especially when both activations and network weights are binary  \cite{courbariaux2015binaryconnect, courbariaux2016binarized, rastegari2016xnor}. In this case, the binary convolution can be efficiently implemented using the XNOR gate, resulting in model compression ratio of $\sim 32\times$ and speed-up of $\sim 58\times$ on CPU~\cite{rastegari2016xnor}. As there is no such thing as a free lunch, these impressive figures come at the cost of reduced accuracy. For example, there is $ \sim 18 \%$ drop in top-1 accuracy between a real-valued ResNet-18 and its binary counterpart on ImageNet~\cite{rastegari2016xnor}. The main aim of this work is to try to bridge this gap by training more powerful binary networks.    

The key observation made in the seminal work of~\cite{rastegari2016xnor} is that one can compensate to some extent for the error caused by the binary approximation by re-scaling the output of the binary convolution using real-valued scale factors. This maintains all the advantages of binary convolutions by adding a negligible number of parameters and complexity. Our key observation in this work is regarding how these scaling factors are computed. While the authors of~\cite{rastegari2016xnor} used an analytic approximation for each layer, we argue that this is sub-optimal for learning the task in hand, and propose to learn these factors discriminatively via backpropagation. This allows the network to optimize the scaling factors with respect to the loss function of the task in hand rather than trying to reduce the approximation error induced by the binarization.

In particular, we make the following \textbf{contributions}: 
\begin{itemize}
    \item 
    We propose to fuse the activation and weight scaling factors into a single one that is learned discriminatively via backpropagation. Further, we propose several ways to construct the shape of these scale factors keeping the complexity at test time fixed. Our constructs increase the expressivity of the scaling factors that are now statistically learned both spatially and channel-wise (Section~\ref{ssec:weight_feat_binarization}).
    \item
    We empirically measure the accuracy of our approximations and show that they are significantly more accurate than the analytically calculated one (Section~\ref{ssec:emperical-performance}).
    \item 
    We show that our improved training of binary networks is agnostic to network architecture used by applying it to both shallow and deep residual networks. 
     \item 
    Exhaustive experiments conducted on the challenging ImageNet dataset show that our method offers an improvement of more than 6\% in absolute terms over the state-of-the-art (Section~\ref{sec:results}). 
\end{itemize}

\section{Related work}\label{sec:related_work}

This section offers a brief overview of the relevant work on designing deep learning methods suitable for running under tight computational constraints. In order to achieve this, in the recent years, a series of different techniques have been proposed such as: network pruning~\cite{lin2017runtime,han2015deep,molchanov2016pruning}, which consists of removing the least important weights/activations, conditional computation~\cite{bengio2015conditional}, low rank approximations~\cite{kim2015compression,lebedev2014speeding,kossaifi2019t}, which decompose the weights and enforce a low rank constraint on them, designing of efficient architectures~\cite{he2016deep,he2016identity} and network quantization~\cite{zhou2016dorefa,lin2015fixed}. A detailed review of all of these different techniques goes beyond the scope of this paper, so herein we will focus on presenting the closest to our work: designing efficient architectures and network quantization, focusing, more specifically, on network binarization~\cite{courbariaux2016binarized,rastegari2016xnor,bulat2017binarized}.

\subsection{Efficient neural networks}\label{ssec:efficient-nn}

With the increase in popularity of mobile devices a large body of work on designing efficient convolutional networks has emerged. From an architectural standpoint such methods take either a holistic approach (i.e. improving the overall structure) or local, by improving the convolutional block or the convolution operation itself.

\textbf{Local optimization}. Since the groundbreaking work of Krizhevsky et al.~\cite{krizhevsky2012imagenet} that introduced the AlexNet architecture, subsequent work attempted to improve the overall accuracy while reducing the computational demand. In VGG~\cite{simonyan2014very}, the large convolutional filters previously used in AlexNet (up to $11\times 11$) were replaced with a series of smaller $3\times 3$ filters that had an equivalent receptive field size (e.g. 2 convolutional layers with $3\times 3$ filters are equivalent with one that has $5\times 5$ filters). This idea is further explored in the numerous versions of the inception block~\cite{szegedy2015going,szegedy2016rethinking,szegedy2017inception} where one of the proposed changes was to decompose some of the convolutional layers that have a $3 \times 3$ kernel into two consecutive layers with $1 \times 3$ and $3 \times 1$ kernels, respectively. In~\cite{he2016deep}, He et al. introduce the so-called ``bottleneck'' block that reduces the number of channels processed by large filters (i.e. $3 \times 3$) using 2 convolutional layers with a $1 \times 1$ kernel that project the features into a lower dimensional space and back. This idea is further explored in~\cite{xie2016aggregated} where the convolutional layers with $3\times3$ filters are decomposed into a series of independent smaller layers with the help of grouped convolutions. Similar ideas are explored in MobileNet~\cite{howard2017mobilenets} and MobileNet-v2~\cite{sandler2018mobilenetv2} that use point-wise grouped convolutions and inverted residual modules, respectively. Chen et al.~\cite{chen2019drop} propose to factorize the feature maps of a convolutional network by their frequencies, introducing an adapted convolution operation that stores and processes feature maps that vary spatially ``slower'' at a lower spatial resolution reducing the overall computation cost and memory footprint.

\textbf{Holistic optimization}. Most of the recent architectures build upon the landmark work of He et al.~\cite{he2016deep} that proposed the so-called Residual networks. Dense-Net~\cite{huang2016densely}, for example, adds a connection from a given layer inside a macro-module to every other layer. In~\cite{redmon2016you} and its improved versions~\cite{redmon2017yolo9000,redmon2018yolov3}, the authors introduce the ``You Only Look Once''(YOLO) architecture which designs a new framework for object detection with a specially optimized topology for the network backbone that allows for real-time or near real-time performance on a modern high-end GPU.

Note, that in this work, we do not attempt to improve the network architecture itself and instead explore our novel approach in the context of both shallow (AlexNet~\cite{krizhevsky2012imagenet}) and deep residual networks (ResNets~\cite{he2016deep,he2016identity}) showing that our method is orthogonal and complementary to methods that propose better network topologies.

\subsection{Network binarization}\label{ssec:network_binarization}

With the rise of in-hardware support for low-precision operations, recently, network quantization has emerged as a natural way of improving the efficiency of CNNs by aligning them with the underlining hardware implementations. Of particular interest is the extreme case of quantization - network binarization, where the features and the weights of a neural network are quantized to two states, typically $\{\pm 1\}$.

While initially binarization was thought to be unfeasible due to the extreme quantization errors introduced, recent work suggests otherwise~\cite{courbariaux2014training,courbariaux2015binaryconnect,courbariaux2016binarized,rastegari2016xnor}. However, despite of the recent effort, training fully binary networks remains notoriously difficult. It is important to note that among the methods that make use of binarization, some of them binarize the weights~\cite{courbariaux2015binaryconnect,faraone2018syq,tang2017train} while keeping the input signal either real or quantized to n-bits, and some of them additionally binarize the signal too~\cite{rastegari2016xnor,courbariaux2016binarized,bulat2017binarized,bulat2018hierarchical}. Because the input features dominate the overall memory consumption (especially for large batch sizes), an effective binarization approach should ideally binarize both weights and features. Not only does this reduce the memory footprint, but also allows the replacement of all the multiplications used in a convolutional layer with bitwise operations. In this work, we study and attempt to improve this particular case of interest.

The method of~\cite{zhou2016dorefa} Zhou et al. allocates a different number of bits per each network component based on their sensitivity to numerical inaccuracies. As such, the method proposes to use 1 bit for the weights, 2 for the activations and 6 for the gradients. \cite{wang2018two}~introduces a n-bit quantization method ($n\geq2$), in which a low-bit code is firstly composed and then a transformation function is learned. In~\cite{faraone2018syq}, the authors quantize the weights using 1 to 2 bits and the features using 2 to 8 bits, by learning a symmetric codebook for each particular weight subgroup. 

The foundations of the fully binarized networks were laid out in~\cite{courbariaux2014training} and the follow-up works of~\cite{courbariaux2015binaryconnect, courbariaux2016binarized}. To reduce the quantization error and improve their expressivity, in~\cite{rastegari2016xnor}, the authors propose to use two real-valued scaling factors, one for the weights and one for activations. The proposed XNOR-Net~\cite{rastegari2016xnor} is the first method to report good results on a large-scale dataset (ImageNet). In this work, we propose to fuse the activation and weight scaling factors into a single one which is learned discriminatively instead of computing them analytically as in~\cite{rastegari2016xnor}. We also motivate and explore various ways for constructing the shape of the factors.

Bulat\&Tzimiropoulos~\cite{bulat2017binarized} propose a novel residual block specifically designed for binary networks for localization tasks, addressing the binarization problem from a network topology standpoint. In~\cite{zhou2018explicit}, Zhou et al. proposes a loss-aware binarization method that jointly regularizes the approximation error and the task loss.  Motivated by the fact that for large batch sizes most of the memory is taken by activations, the method of~\cite{mishra2017wrpn} proposes to increase the network width (represented by the number of channels of a given convolutional layer). 
Similarly, the work of~\cite{lin2017towards} introduces the ABC-Net that uses up to 5 parallel binary convolutional layers to approximate a real one. While this increases the network accuracy, it does so at a high cost as the resulting network is up to 5$\times$ slower. In contrast, we improve the overall accuracy of fully binarized networks within the same computational budget.

\section{Background}\label{sec:background}

This section reviews the binarization process proposed in~\cite{courbariaux2015binaryconnect} alongside XNOR-Net, its improved version from~\cite{rastegari2016xnor}, which still represents the state-of-the-art method for training binary networks.

For a given layer $L$ of a CNN architecture we denote with $\mytensor{W}\in \mathbb{R}^{o\times c\times w \times h}$ and $\mytensor{I}\in \mathbb{R}^{c\times w_{in} \times h_{in}}$ its weights and input features, where $o$ represents the number of output channels, $c$ the number of input channels and ($w,h$) the width and height of the convolutional kernel. Moreover, $w_{in}\geq w$ and $h_{in}\geq h$ represent the spatial dimensions of $\mytensor{I}$. Following~\cite{courbariaux2015binaryconnect}, the binarization is done by taking the sign of the weights and input features, where

\begin{equation}\label{eq:sign}
    \text{sign}(x) = \begin{cases} -1, & \mbox{if } x\leq 0 \\ 1, & \mbox{if } x>0 \end{cases}.
\end{equation}

Because binarization is a highly destructive process in which large quantization errors are induced, the achieved accuracy, especially on challenging datasets (such as Imagenet) is low. To alleviate this, Rastegari et al.~\cite{rastegari2016xnor} introduce two analytically calculated real-valued scaling factors, one for the weights and one for the input features, which are used to re-weight the output of a binary convolution as follows:

\begin{equation}\label{eq:baseline}
		\mytensor{I} \ast \mytensor{W} \approx \left(\text{sign}(\mytensor{I}) \oast \text{sign}(\mytensor{W}) \right) \odot \mytensor{K} \myvector{\alpha},
\end{equation}
\noindent where $\odot$ denotes the element-wise multiplication, $\ast$ the real-valued convolution operation and $\oast$ its binary counterpart (implemented using bitwise operations), $\myvector{\alpha}_{i} = \frac{\norm{\mytensor{W}_{i, \mycolon, \mycolon, \mycolon}}_{\ell 1}}{n},~i=\{1, 2, \cdots, o\}$, $n = c\times w \times h$ is the weight scaling factor, and $\mytensor{K}$ the activation scaling factor. $\mytensor{K}$ is efficiently computed by convolving $\mytensor{A} = \frac{\sum \lVert I_{i,\mycolon,\mycolon}\rVert}{c}$ with a 2D filter $k \in \mathbb{R}^{w\times h}$, where $\forall{ij}~k_{ij}=\frac{1}{w\times h}$. Note, that as the calculation of $\mytensor{K}$ is relatively expensive due to the fact that it is recomputed at each forward pass, it is common to drop it at the expense of a slight drop in accuracy~\cite{rastegari2016xnor,bulat2017binarized}. In contrast, in this work, we fuse $\myvector{\alpha}$ and  $\mytensor{K}$ into a single factor that is learned via backpropagation. In the process, we motivate and explore various ways of forming the shape of this factor.

\section{Method}\label{sec:method}

In this section, we firstly present, in Sub-section~\ref{ssec:weight_feat_binarization}, our proposed improved binarization technique, coined XNOR-Net++, that increases the representational power of binary networks by describing novel ways to construct scaling factors for the binary convolutional layers. Next, Sub-sections~\ref{ssec:emperical-performance} and~\ref{ssec:efficiency-analysis} analyze empirically the performance of our method and, the speed-up and memory savings offered.

\subsection{XNOR-Net++}~\label{ssec:weight_feat_binarization}

As mentioned earlier (see Section~\ref{ssec:emperical-performance}), directly binarizing the weights and the input features of a given layer using the sign function is known to induce high quantization errors. To alleviate this, in~\cite{rastegari2016xnor}, one of the key elements that allowed the training of more accurate binary networks on challenging datasets was the introduction of the scaling factors $\myvector{\alpha}$ and $\mytensor{K}$ for the weight and features, respectively (see also Section~\ref{sec:background}).
While the analytical solution provided in~\cite{rastegari2016xnor} works well, in general, it fails to take in consideration the overall task at hand and has limited flexibility since obtaining a good minimum is directly tight with the distribution of the binary weights. Moreover, computing the scaling factor with respect to the features is relatively expensive and needs to be done for every new input.  

To alleviate this, in this work, we propose to fuse the activation and weight scaling factors into a single one, denoted as $\Gamma$, that is learned discriminatively via backpropagation. This allows us to capture a statistical representation of our data, facilitates the learning process and even has the advantage that at test time the analytic calculation of these factors is not required, thus reducing the number of real-valued operations. In particular, we propose to re-formulate Eq.~(\ref{eq:baseline}) as:

\begin{equation}\label{eq:ours}
	\mytensor{I} \ast \mytensor{W} \approx \left(\text{sign}(\mytensor{I}) \oast \text{sign}(\mytensor{W}) \right) \odot \Gamma 
\end{equation}

This new formulation allows us to explore various ways of constructing the shape of $\Gamma$ during training. Specifically, we propose to construct $\Gamma$ in the following 4 ways:

\noindent\textbf{Case 1:} \begin{equation}\label{eq:alpha}
	\Gamma = \myvector{\alpha},~~\myvector{\alpha} \in \mathbb{R}^{o \times 1 \times 1}
\end{equation}

\noindent\textbf{Case 2:} 
\begin{equation}\label{eq:alpha-big}
	\Gamma = \myvector{\alpha},~~\myvector{\alpha} \in \mathbb{R}^{o\times h_{out} \times w_{out}}
\end{equation}

\noindent\textbf{Case 3:} 
\begin{equation}\label{eq:alpha-beta}
	\Gamma = \myvector{\alpha} \otimes \myvector{\beta},~~ \myvector{\alpha} \in \mathbb{R}^{o}, \myvector{\beta} \in \mathbb{R}^{h_{out} \times w_{out}}
\end{equation}

\noindent\textbf{Case 4:} 
\begin{equation}\label{eq:alpha-beta-gamma}
	\Gamma = \myvector{\alpha} \otimes \myvector{\beta} \otimes \myvector{\gamma},~~ \myvector{\alpha} \in \mathbb{R}^{o}, \myvector{\beta} \in \mathbb{R}^{h_{out}}, \myvector{\gamma} \in \mathbb{R}^{w_{out}}
\end{equation}

\noindent\textbf{Case 1:} The first straightforward approach is to simply learn one scaling factor per each input channel, similarly to what a Batch Normalization layer would do. As Tables~\ref{tab:results-alpha} and~\ref{tab:results-emperical} show learning this alone instead of computing it analytically is already significantly better than the analytically calculated factors proposed in~\cite{rastegari2016xnor}.

\noindent\textbf{Case 2:} While Case 1 works reasonably well, it fails to capture the information en-capsuled over the spatial dimensions. To address this, in Eq.~(\ref{eq:alpha-big}), we propose to learn a dense scaling, one value for each output pixel, which as Table~\ref{tab:results-alpha} shows, performs 0.6\% better than the previous version.

\noindent\textbf{Case 3:} While Case 2 performs 0.6\% better than the previous version, it is relatively large in size and harder to optimize often leading to some sort of overfitting. As such, in Eq.~(\ref{eq:alpha-beta}), we propose to decompose this dense scaling into two terms combined using an outer product. As Eq.~(\ref{eq:alpha-beta}) shows, $\myvector{\alpha}$ learns the statistics over the output channel dimension while $\myvector{\beta}$ over the spatial dimensions. With a reduced number of parameters, when compared to our previous version, this further boosts the performance by 0.6\%. 

\noindent\textbf{Case 4:} Upon analyzing $\myvector{\beta}$ in Case 3, we noticed that it is low rank and as such further compressions are possible. This leads us to the final version shown in Eq.~(\ref{eq:alpha-beta-gamma}), where we learn a rank-1 factor for each mode (channels, height, weight). This further reduces the number of parameters making their number negligible when compared with the overall number of weights present inside a layer and improves the performance by a further 0.4\%. 

Note, that in all cases, during testing, the factors are merged together into a single one and a single element-wise product takes place (see also Section~\ref{ssec:efficiency-analysis}).

\begin{table}[!htbp]
	\begin{center}
		\begin{tabular}{|l|c|c|c|c|}
			\hline
			Method & shapes
			  & Top-1 acc. & Top-5 acc. \\
			\hline\hline
			baseline~\cite{rastegari2016xnor} & - & 51.2\% & 73.2\%  \\
			\hline \hline
			  \textbf{Case 1:} $\myvector{\alpha}$ & $\myvector{\alpha} \in \mathbb{R}^{o \times 1 \times 1}$ & 55.5\% & 78.5\%  \\
			  \hline
			  \textbf{Case 2:} $\myvector{\alpha}$ &$\myvector{\alpha} \in \mathbb{R}^{o\times h_{out} \times w_{out}}$ &  56.1\%& 79.0\% \\
			  \hline
			   \textbf{Case 3:} $\myvector{\alpha \otimes \beta}$ & $\myvector{\alpha} \in \mathbb{R}^{o}, \myvector{\beta} \in \mathbb{R}^{w_{out} \times h_{out}} $  & 56.7\% & 79.5\%\\
			  \hline
			   \textbf{Case 4:} $\myvector{\alpha \otimes \beta \otimes \gamma}$ & \begin{tabular}{@{}c@{}}$\myvector{\alpha} \in \mathbb{R}^{o}, \myvector{\beta} \in \mathbb{R}^{w_{out}} $ \\ $\myvector{\gamma} \in \mathbb{R}^{h_{out}}$\end{tabular}  & \textbf{57.1}\% & \textbf{79.9}\%
	
			  \\			  
			\hline
		\end{tabular}
	\end{center}
	\caption{Top-1 and Top-5 classification accuracy using a binarized ResNet-18 on Imagenet for various ways of constructing the scaling factor. $\myvector{\alpha}, \myvector{\beta}, \myvector{\gamma}$ are statistically learned via backpropagation. Note that, at test time, all of them can be merged into a single factor, and a single element-wise multiplication is required. }
	\label{tab:results-alpha}
\end{table}

\subsection{Empirical performance analysis}\label{ssec:emperical-performance}

In Section~\ref{sec:results}, we showcased the advantages of learning a single scaling factor $\Gamma$ discriminatively and explored various ways to construct it for further improving the achieved accuracy for the task of ImageNet classification. Herein, we reach similar conclusions by analyzing the quantization loss, and more specifically, by showing that, for a given real-valued convolutional layer, our method can approximate its output with a binary convolution with higher fidelity. 

For our experiments, we created a convolutional layer with $\mytensor{W} \in \mathbb{R}^{64\times 64 \times 3 \times 3}$ and $\mytensor{I} \in \mathbb{R}^{64\times16\times16}$ both initialized from a normal distribution. We then tried to compute an equivalent binary layer having as target to minimize the reconstruction error between its output and the one of the real-valued layer. As shown in~\cite{rastegari2016xnor}, the optimal solution for the binary weights is given by $\text{sign}(\mytensor{W})$. Although~\cite{rastegari2016xnor} does develop an analytic solution for the weights, this solution is only an approximation. Hence, for the needs of our experiment, we fixed the binary weights as $\text{sign}(\mytensor{W})$ and then trained the scaling factors for the cases proposed in section~\ref{ssec:weight_feat_binarization}. The layer is trained until convergence using Adam~\cite{kingma2014adam} (SGD and RMSProp gave similar results). We then repeated the process 100 times, for different $\mytensor{W}$ and $\mytensor{I}$. The L1 distance between the output of the real convolution and that of the binary one is shown in Table~\ref{tab:results-emperical}. Notice that our methods consistently outperforms~\cite{rastegari2016xnor} by a large margin.

\begin{table}[!htbp]
	\begin{center}
		\begin{tabular}{|l|c|c|c|}
			\hline
			Method & shapes
			  & L1 distance \\
			\hline \hline 
			Direct binarization~\cite{courbariaux2016binarized} & - & $6.36\pm0.04$ \\
			Baseline XNOR~\cite{rastegari2016xnor} & - & $0.095 \pm0.002$  \\
			\hline \hline 
			  \textbf{Case 1:} $\myvector{\alpha}$ & $\myvector{\alpha} \in \mathbb{R}^{o\times 1 \times 1}$ & $0.038\pm0.001$  \\
			  \hline
			  \textbf{Case 3:} $\myvector{\alpha \otimes \beta}$ & $\myvector{\alpha} \in \mathbb{R}^{o}, \myvector{\beta} \in \mathbb{R}^{w_{out} \times h_{out}} $ & $0.037\pm0.001$ \\
			  \hline
			  \textbf{Case 4:} $\myvector{\alpha \otimes \beta \otimes \gamma}$ & \begin{tabular}{@{}c@{}}$\myvector{\alpha} \in \mathbb{R}^{o}, \myvector{\beta} \in \mathbb{R}^{w_{out}} $ \\ $\myvector{\gamma} \in \mathbb{R}^{h_{out}}$\end{tabular} & $\mathbf{0.035\pm0.001}$ \\			  
			\hline
		\end{tabular}
	\end{center}
	\caption{L1 distance between the output of a real-valued convolutional layer and its binary counterpart using different methods for learning the scale factors. In all cases, the binary weights are fixed as $\text{sign}(\mytensor{W})$. Note that, in XNOR-Net, the scale factors are not learned but analytically calculated.}
	\label{tab:results-emperical}
\end{table}

\subsection{Efficiency analysis}\label{ssec:efficiency-analysis}

\begin{figure}
	\centering
	\subfigure[]{\label{fig:a}\includegraphics[width=45mm]{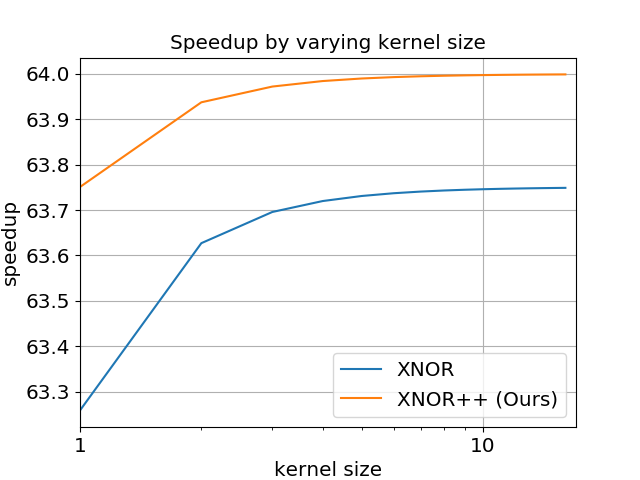}}
	\subfigure[]{\label{fig:b}\includegraphics[width=45mm]{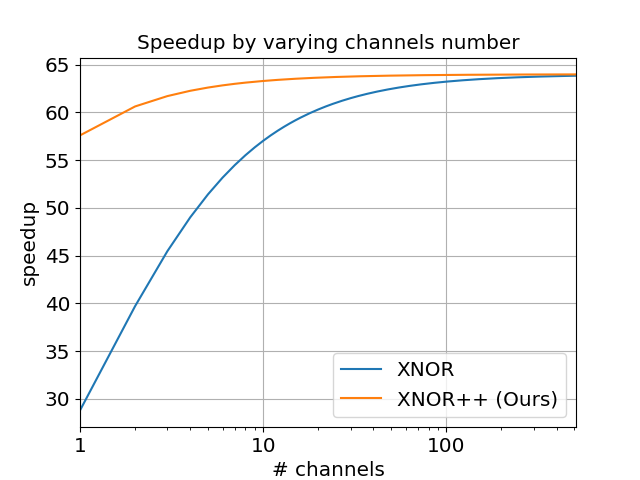}}
	\caption{Theoretical speed-up offered by our method and~\cite{rastegari2016xnor}.}
	\label{fig:speed-comparison}
	\end{figure}

An important aspect of binary convolutions are the speed-ups offered. Assuming an implementation with no algorithmic optimizations, the total number of operations for a given convolutional layer is $N=c\times w \times h \times w_{out} \times h_{out} \times o$. Given the usage of bit-packing and an SIMD approach, a modern CPU can execute 64$\times$ more binary operations per clock than multiplications. Since the XNOR-Net~\cite{rastegari2016xnor} method computes an independent scale for the weights and  the features, in addition to $N$ XNOR ops, the binary layer will require $2\times c \times h_{out} \times w_{out}$ multiplications and $c\times h_{out} \times w_{out} \times h \times w$ additions, making the overall theoretical speed-up approx. equal to:

\begin{equation}
	S_{XNOR} = \frac{64 \times w \times h \times o}{ w \times h \times o + 2  +  h \times w}.
\end{equation}

In contrast, since our method fuses the scaling factors, it only requires $c \times w_{out} \times h_{out}$ additional floating point operations:

\begin{equation}
	S_{OURS} = \frac{64 \times w \times h \times o}{ w \times h \times o + 1}.
\end{equation}

Notice that the speed-up is independent of the input feature resolution and does not include the memory access cost. Assuming a layer with 256 output channels and a kernel size of $3\times3$ (one of the most common layers found in a Resnet architecture~\cite{he2016deep}), $S_{XNOR} \approx 63.69\times$ while $S_{OURS} \approx 63.98\times$. In terms of storage, similarly to BNN and XNOR-Net, our method can take advantage of bit-packing offering a space saving of $\approx 64\times$.

\section{Results}\label{sec:results}

In this section, we describe the experimental setting used in our work and compare our method against other state-of-the-art binary networks. We show that our approach largely outperforms the current top performing methods by more than 6\% on ImageNet classification.

\subsection{Experimental setup}

This section describes the experimental setup of our paper going through the dataset and networks used and providing details regarding the training process.

\subsubsection{Network architecture}
Herein, we describe the topology of the two networks used: AlexNet~\cite{krizhevsky2012imagenet} and ResNet-18~\cite{he2016deep} alongside their modifications, if any.

\noindent\textbf{ResNet-18.} We preserved the overall network architecture (i.e. 18 layers distributed over 4 macro-blocks; except for the first and last layer all of them are grouped in pairs of 2 inside a basic block~\cite{he2016deep}). We note that we followed~\cite{rastegari2016xnor} and used the basic block version with pre-activation~\cite{he2016identity} moving the activation function after the convolution and adding a sign function before it.

\noindent\textbf{AlexNet.} In line with previous works~\cite{rastegari2016xnor,courbariaux2016binarized}, we  removed  the local normalization operation and added a batch normalization~\cite{ioffe2015batch} layer followed by a sign activation before each convolutional layer. Additionally, we kept the dropout on both fully connected layers setting its value to 0.5. 

As in~\cite{rastegari2016xnor}, the first and last layers for both networks were kept real.

\subsubsection{Datasets} We trained and evaluated our models on ImageNet~\cite{deng2009imagenet}.  ImageNet is a large-scale image recognition dataset containing 1.2M training and 50,000 validation samples distributed over 1000 non-overlapping classes. 

\subsubsection{Training}
For training both ResNet-18~\cite{he2016deep} and AlexNet~\cite{krizhevsky2012imagenet} we follow the common practices used for training binary nets~\cite{rastegari2016xnor}: we resized the input images to $256\times 256$px and then randomly cropped them during training to $224\times 224$px for ResNet and $227\times 227$px for AlexNet, while during testing we center-cropped them to the corresponding sizes. For both models, the initial learning rate was set to $10^{-3}$ and the weight decay to $10^{-5}$. The learning rate was dropped during training every 25 epochs by a factor of 10. The entire training process runs for 80 epochs. Similarly to~\cite{rastegari2016xnor}, we used a batch size of 400 for AlexNet and 256 for ResNet. The weights are initialized as in~\cite{he2016deep}.

All of our models were trained using Adam~\cite{kingma2014adam}. They are implemented in Pytorch~\cite{paszke2017automatic}.

\subsection{Comparison with state-of-the-art}\label{ssec:comparison-sota}

In this section, we compare the performance of our approach against those of other state-of-the-art methods that binarize both the weights and the features within the same computational budget. We note that most of prior work only binarize the weights and use either full precision or n-bits quantized activations and as such cannot take advantage of the large speed-ups offered by full binary convolutions. We also note that to allow for a fair comparison, we compare only against methods that use the same number of weights: to achieve high accuracy, ABC-Net increases the network size $25 \times$, while their version which has the same number of parameters as ours (i.e. for M=N=1 using a ResNet-18, where M and N represent the expansion rates for the features and weights respectively) achieves a top-1 accuracy of 42.2\% only (vs 57.1\% achieved by our methods).

Our results are summarized in Table~\ref{tab:results-sota}: when using ResNet-18, our method significantly outperforms the state-of-the-art by about 6\% in terms of absolute error using both Top-1 and Top-5 metrics. For AlexNet, we observe that the improvement was not as great. In general, we found that AlexNet was much harder to train and prone to overfitting.

\begin{table}[!htbp]
	\begin{center}
		\begin{tabular}{|l|c|c|c|c|c|}
			\hline
			\multicolumn{1}{|c|}{Method} & \multicolumn{2}{c|}{AlexNet} & \multicolumn{2}{c|}{ResNet-18}\\
			\cline{2-5}
			 & Top-1 accuracy & Top-5 accuracy & Top-1 accuracy & Top-5 accuracy \\
			\hline\hline
			  BNN~\cite{courbariaux2016binarized} & 41.8\% & 67.1\% & 42.2\% & 69.2\%  \\
			  XNOR-Net~\cite{rastegari2016xnor} & 44.2\% & 69.2\% & 51.2\%& 73.2\% \\
			  Bethge et al.~\cite{bethge2018training} & - & - & 54.4\%& 77.5\% \\
			  \textbf{Ours} & \textbf{46.9}\% & \textbf{71.0}\% & \textbf{57.1}\% & \textbf{79.9}\%\\
			  \hline
			  Real valued~\cite{krizhevsky2012imagenet} & 56.6\% & 80.2\% & 69.3\%& 89.2\%\\
			\hline
		\end{tabular}
	\end{center}
	\caption{Top-1 and Top-5 classification accuracy using binarized AlexNet and ResNet-18 architectures on the validation set of Imagenet.}
	\label{tab:results-sota}
\end{table}

\section{Conclusion}\label{sec:conclusion}

We revisited the calculation of scale factors used to re-weight the output of binary convolutions by proposing to learn them discriminatively via backpropagation. We also explored different shapes for these factors. We showed large improvements of up to $6\%$ on ImageNet classification using ResNet-18.

\bibliography{xnornet_plus}
\end{document}